# Data Dependent Kernel Approximation using Pseudo Random Fourier Features

Bharath Bhushan Damodaran[1], Nicolas Courty[1], and Philippe-Henri Gosselin[2]

[1] IRISA, Univ. Bretagne-Sud, Vannes
[2] ETIS UMR 8051, Univ. Paris Seine, Univ. Cergy-Pontoise, ENSEA, CNRS
{bharath-bhushan.damodaran, nicolas.courty}@irisa.fr,
philippe-henri.gosselin@ensea.fr

**Abstract.** Kernel methods are powerful and flexible approach to solve many problems in machine learning. Due to the pairwise evaluations in kernel methods, the complexity of kernel computation grows as the data size increases; thus the applicability of kernel methods is limited for large scale datasets. Random Fourier Features (RFF) has been proposed to scale the kernel method for solving large scale datasets by approximating kernel function using randomized Fourier features. While this method proved very popular, still it exists shortcomings to be effectively used. As RFF samples the randomized features from a distribution independent of training data, it requires sufficient large number of feature expansions to have similar performances to kernelized classifiers, and this is proportional to the number samples in the dataset. Thus, reducing the number of feature dimensions is necessary to effectively scale to large datasets. In this paper, we propose a kernel approximation method in a data dependent way, coined as Pseudo Random Fourier Features (PRFF) for reducing the number of feature dimensions and also to improve the prediction performance. The proposed approach is evaluated on classification and regression problems and compared with the RFF, orthogonal random features and Nyström approach.

## 1 Introduction

Kernel methods is an efficient tool to capture the non-linear dependencies in the data. It is achieved by implicitly mapping the input data into the infinite dimension feature space, thus allowing the non-linear data representations [1]. This representation comes with the cost involving the computational complexity in terms of memory and time, and it quadratically grows with data size. Furthermore, in the prediction phase of learning task it is necessary to store the large parts of training dataset. This behaviour limits the potential of kernel methods for exploiting the large scale datasets.

Kernel approximation technique is considered as the alternative way to harness the expressive power of the kernel methods and to scale up for the large scale data: they find an explicit mapping function to transform the data into a new Hilbert space where the dot product approximates the kernel function [2].

Following the seminal work in [2], random Fourier features (RFF) has gained significant attention to scale up kernel machines. RFF is a simple and effective approach, where the kernel function is expressed as the Fourier transform of a probability measure. As the RFF samples the coefficients from the distribution independently of training data, it requires large number of feature dimensions for better approximation quality. In order to overcome this shortcoming of RFF, reducing number of feature dimension becomes essential to scale for large datasets. The objective of this paper is to develop a kernel approximation method in a data dependent fashion, coined as pseudo random Fourier features (PRFF) such that kernel approximation can be achieved using smaller number of features and also to increase the prediction performance of the learning task.

**Related works:** Following the line of reducing the number of required feature dimensions of RFF, several works theoretically and empirically studied the RFF to better understand the required number of features to approximate the gram matrix [3–6]. The approximation quality of RFF entirely depends on the Monte-Carlo sampling, and thus the approximation quality can be improved by clever sampling strategies. Yang *et.al* [7] used the quasi-Monte Carlo approximation using low-discrepancy sequence of points instead of random samples. Hamid *et. al* [8] oversampled the coefficient vectors and projected into a lower dimensional subspace. Yu *et.al* [9] optimized the features relative to the classification accuracy in a data dependent fashion, while our proposed method optimizes the features relative to the approximation error and have the advantage of being unsupervised. Furthermore, our approach to learn the feature coefficients are also different. More recently, Yu *et.al* [10] proposed the use of scaled orthogonal random matrix instead of random Gaussian matrix for kernel approximation and they showed that their method has lower kernel approximation error compared to the existing methods. Besides that, the gram matrix was also approximated by the low rank approximation methods such as the Nyström method [11, 12]. The approximation quality of Nyström method is better than RFF, especially when there is a large eigen gap present in the data [5]. Related to this, Mukuta *et. al* [13] approximates the gram matrix by solving the eigen function decomposition using the distribution estimated from the training data, and their performance is similar to Nyström method.

## 2 Random Fourier Features

We begin by shortly describing about the random Fourier features, let $k$ be a positive definite and shift invariant kernel defined on $\mathcal{R}^d$, $k(\mathbf{x}-\mathbf{y}) = k(\mathbf{x},\mathbf{y})$. According to the Bochner's theorem, every positive definite, continuous and shift invariant kernel $k$ is the Fourier transform of the non-negative measure [14]. Let $p(\mathbf{w})$ be the Fourier transform of $k$, then

$$k(\mathbf{x},\mathbf{y}) = k(\mathbf{x}-\mathbf{y}) = \int_{\mathcal{R}^d} p(\mathbf{w}) e^{i\mathbf{w}^t(\mathbf{x}-\mathbf{y})} d\mathbf{w}, \tag{1}$$

since the probability measure $p(\mathbf{w})$ and $k$ are real, the eq. (1) can be approximated using Monte-Carlo sampling as:

$$\hat{k}(\mathbf{x},\mathbf{y}) \simeq \frac{2}{D}\sum_{i=1}^{D} cos(\mathbf{w}_i^t\mathbf{x}+b_i)cos(\mathbf{w}_i^t\mathbf{y}+b_i) = \phi(\mathbf{x})^t\phi(\mathbf{y}) \tag{2}$$

where $\phi(\mathbf{x}) = \sqrt{\frac{2}{D}}\left[cos(\mathbf{w}_1^t\mathbf{x}+b_1), cos(\mathbf{w}_2^t\mathbf{x}+b_2), \ldots, cos(\mathbf{w}_D^t\mathbf{x}+b_D)\right]$ is called Random Fourier Features. The coefficient vector $\mathbf{w}$'s are sampled from the Gaussian distribution as $N(0,\sigma^{-1})$, where $\sigma$ is the band width of Gaussian RBF kernel and the bias $b$'s are sampled from the uniform distribution on $U(0,2\pi)$. Now, the non-linear learning using RFF is performed as: (i) draw the $\mathbf{w}$'s from $N(0,\sigma^{-1})$ and $b$ from $U(0,2\pi)$, (ii) compute the random Fourier features $\phi$ of the input data, (iii) and perform linear machines on the random Fourier features. Thus, the random Fourier features can scale the kernel machines for the large scale data. Since the $\mathbf{w}$'s are sampled from the distribution independent of the training data, in order to have good prediction performance of learning machine, $D$ should be large enough, and this number is in the order of number of samples in the data.

## 3 Proposed method

Reducing the number of random Fourier features is essential to fully benefit the scalability property of RFF for large datasets. Here we propose the framework to enforce the random Fourier features to be data dependent and we call this approach as pseudo random Fourier features (PRFF). The objective of our approach is to approximate the true kernel $k$ using a small number coefficient vectors as,

$$\hat{k}(\mathbf{x},\mathbf{y}) = \frac{2}{M}\sum_{l=1}^{M} cos(\mathbf{w}_l^t\mathbf{x}+b_l)cos(\mathbf{w}_l^t\mathbf{y}+b_l) \tag{3}$$

where $M \prec D$, and the coefficient vectors $\mathbf{w}$'s should have the capability to approximate kernel matrix using smaller dimensions. With out loss of generality, let us assume that we know the feature expansions or coefficient vectors up to $M-1$, then the coefficient vector $\mathbf{w}_M$ can be obtained by minimizing the following loss function $L(\mathbf{w}_M)$ as

$$\begin{aligned} L(\mathbf{w}_M) &= \frac{1}{2N^2}\sum_i\sum_j\Bigg(k(\mathbf{x_i},\mathbf{x_j}) - \frac{2}{M-1}\sum_{l=1}^{M-1} cos(\mathbf{w}_l^t\mathbf{x}+b_l)cos(\mathbf{w}_l^t\mathbf{y}+b_l) \\ &\qquad - 2\; cos(\mathbf{w}_M^t\mathbf{x}+b_M)cos(\mathbf{w}_M^t\mathbf{y}+b_M)\Bigg)^2 \\ &= \frac{1}{2N^2}\sum_i\sum_j\left(k(\mathbf{x_i},\mathbf{x_j}) - \frac{2}{M}\sum_{l=1}^{M} cos(\mathbf{w}_l^t\mathbf{x}+b_l)cos(\mathbf{w}_l^t\mathbf{y}+b_l)\right)^2 \end{aligned} \tag{4}$$

where $N$ is the number of samples, $k(\mathbf{x_i}, \mathbf{x_j})$ is the pairwise evaluation of the true Gaussian kernel among the samples. The coefficient vector $\mathbf{w}_M$ which minimizes the loss function is called as pseudo random Fourier feature (PRFF). Further more, the loss function (4) can also be regularized using $l_2$ regularization in order to enforce the coefficient vector to have a minimal norm as in eq. (5), and this can be interpreted as constraining the loss function to obtain the weight vectors in the low frequency components of the Fourier spectrum.

$$L(\mathbf{w}_M) = \frac{1}{2N^2} \sum_i \sum_j \left( k(\mathbf{x_i}, \mathbf{x_j}) - \frac{2}{M} \sum_{l=1}^{M} cos(\mathbf{w}_l^t \mathbf{x} + b_l) cos(\mathbf{w}_l^t \mathbf{y} + b_l) \right)^2 + \lambda \sum_p \mathbf{w}_{M_p}^2 \quad (5)$$

Batch stochastic gradient descent method is employed to obtain the coefficients from the eq.(5)[3]. The process of obtaining the PRFF coefficients is summarized in Algorithm 1. We learn the coefficient vector one by one in a sequential way until the desired number of features are obtained. More precisely, for each subset (mini-batch) of samples we learn one coefficient vector $\mathbf{w}$ using stochastic gradient descent, and repeat this procedure until $M$ coefficients were learnt.

**Algorithm 1:** Pseudo Random Fourier Features

```
Input: Input Data, W = []
Parameters: Batch size B, learning rate α, regularization parameter λ, and
  no. of feature expansions M
while j < M do
    for each mini-batch do
        while error ≥ tol do
            compute the gradient ∂L/∂w_j
            w_j ← w_j - α. (λ.w_j + ∂L/∂w_j)
        end
        W(:, j)=w_j
        if j == M then
            break
        end
        j ← j + 1
    end
end
Output: pseudo random Fourier feature coefficients W
```

[3] When minimizing the loss function with respect to both $\mathbf{w}$ and $b$, there was no improvement in the prediction performance when compared to minimizing the loss function only with respect to $\mathbf{w}$, thus $b$ is fixed as the constant

Table 1: Statistics of the datasets used in the experiments

| Task | DATA | #Train | #Test | #valid. | #attrib. | Task | DATA | #Train | #Test | #valid. | #attrib. |
|---|---|---|---|---|---|---|---|---|---|---|---|
| Class. | MNIST | 50000 | 10000 | 10000 | 784 | Class. | ADULT | 29304 | 16281 | 3257 | 123 |
| Reg. | CADATA | 15921 | 2949 | 1770 | 8 | Reg. | CENSUS | 16367 | 2273 | 1819 | 119 |
| Reg. | CPU | 5898 | 819 | 656 | 21 | Reg. | YEARMSD | 417343 | 51630 | 46372 | 90 |

The proposed PRFF method can be interpreted in the stochastic gradient boosting framework. Learning each of the feature dimension (coefficient) can be considered as learning the weak learners. For each of mini-batch samples, a new feature dimension are learned with respect to the error of the whole feature coefficients (dimensions) learned so far, thus it minimizes the loss function error in an iterative fashion. The regularization in the loss function is incorporated to avoid the over-fitting to the mini-batch of samples, and this might be necessary to improve the generalization ability of learned PRFF coefficient for the prediction tasks.

## 4 Empirical Evaluations

In order to evaluate efficiency of our proposed method, we conducted experiments with synthetic and real datasets to show the approximation error of the gram matrices, classification accuracy and mean square error. The orthogonal random features is considered as the main baseline approach, as it is proved to have lower approximation error than other existing data independent kernel approximation approaches [10], and Nyström method is considered as the baseline approach in the data dependent kernel approximation approaches, as the existing method [13] only yielded comparable performance to the Nyström approach. In the PRFF, the initialization of $\mathbf{w}$'s in the batch gradient descent are obtained from sampling the Gaussian distribution as equivalent to the random Fourier features, and the maximum number of SGD iterations is set to 100. The mini-batch size $B$ is set to 128. As the initial vector's are same as RFF, our proposed approach can also be seen as fine-tuning RFF coefficients. For the RFF and Nyström method, we used the implementation available in the scikit-learn library[4] and for the orthogonal random features we used the implementation from revrand package[5].

### 4.1 Kernel Approximation

We firstly conducted experiments on synthetic dataset to evaluate the approximation error of the gram matrices of our proposed method, and compared it to Random Fourier features and Orthogonal Random Features. The synthetic dataset was generated with $N = 10000$ vector samples of dimension $d = 10$ and $d = 100$ using Gaussian distribution with zero mean and identity covariance matrix. The kernel bandwidth parameter ($\sigma$) of the Gaussian kernel was estimated on the $5^{th}$ percentile of the pairwise distances of the samples [15, 16]. The quality of kernel approximation was measured using mean the square error ($\frac{1}{N^2} \parallel K_{true} - K_{approx} \parallel^2$). Number of features expansions (dimensions) are set to $M = 25, 50, 100, 200, 300, 400, 500, 750, 1000$, learning rate is set to $\alpha = 50.0$, and regularization parameter to $\lambda = 0$.

[4] http://scikit-learn.org/stable/modules/kernel_approximation.html
[5] https://github.com/NICTA/revrand/

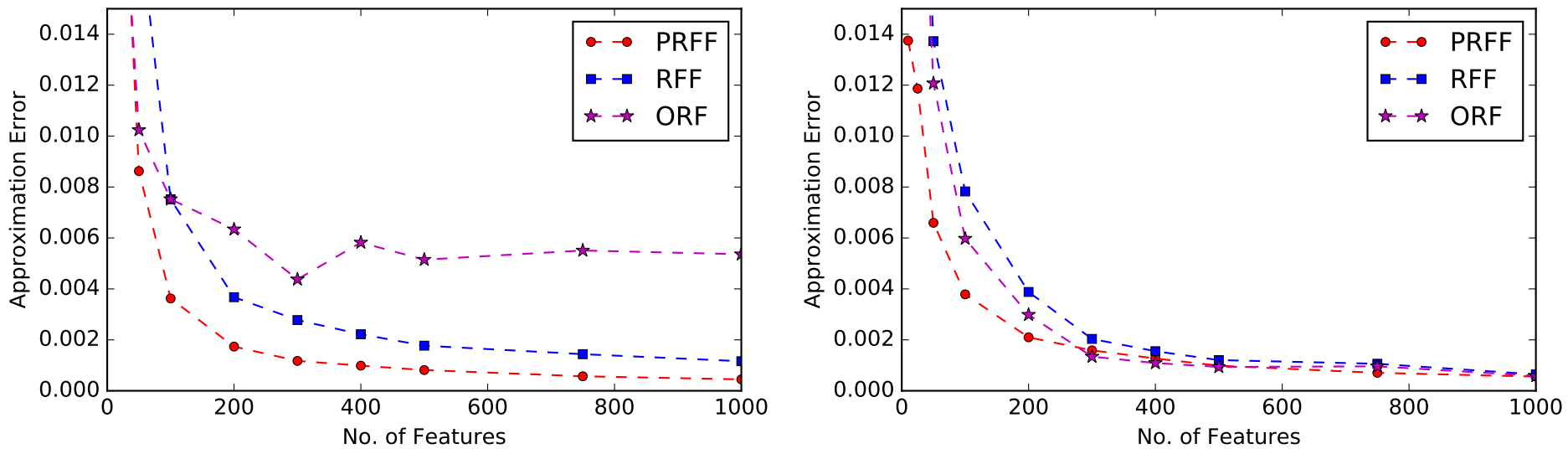

(a) $d = 10$ (b) $d = 100$

Fig. 1: Kernel approximation error using PRFF, RFF, and ORF on the synthetic dataset.

Figure. 1 shows the approximation error for PRFF, RFF and ORF[6]. The figure reveals that the proposed method have better approximation performance than RFF and ORF. The approximation performance of ORF is better in lower dimension than RFF, but it is worse than RFF in higher dimension, especially when $M > 10d$. However when we tested the ORF with the synthetic data generated with $d = 100$, and we didn't observe this behavior (see fig. 1(b)).

We then compared the approximation performance of the proposed method on the real datasets (Table 1). For each dataset, we randomly selected 10000 samples to evaluate the approximation performance. Results are shown in Figure 2. The proposed method yielded the best approximation performance in the lowest dimension in real data as well, and in certain cases the proposed method and ORF are similar. In the Fig. 2 (d) and (e), ORF has similar behaviour to the synthetic dataset with $d = 10$, and in both cases the input dimension of the data is small, thus validating our observation with synthetic dataset.

### 4.2 Classification Accuracy and Mean Square Error

To evaluate the potential of proposed approach in the classification and regression problems we conducted experiments on six real datasets including classification and regression datasets. Table 1 summarizes the statistics of these datasets. Each dataset contains a training and testing sets. We selected 10% of training

[6] For each of the dataset, Nystöm method has a lower approximation error than PRFF, RFF, and ORF

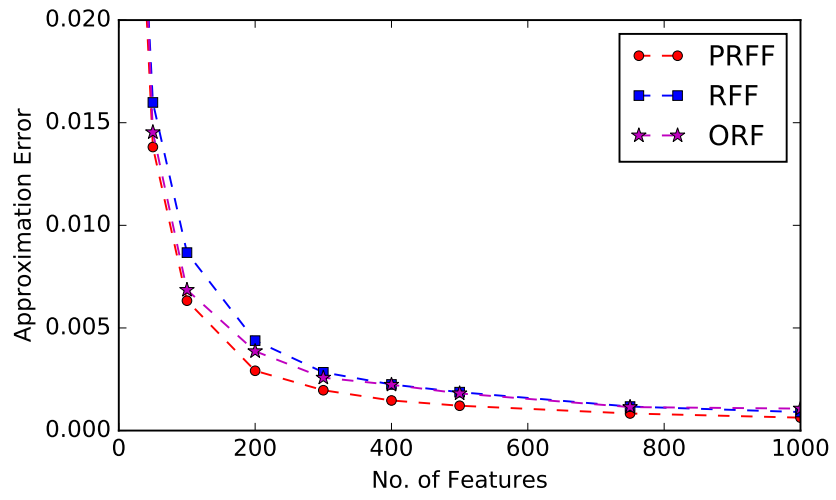

(a) ADULT

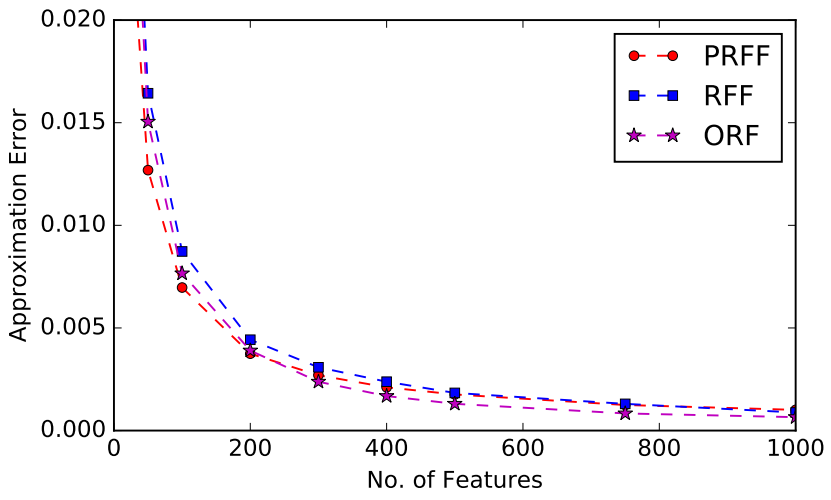

(b) MNIST

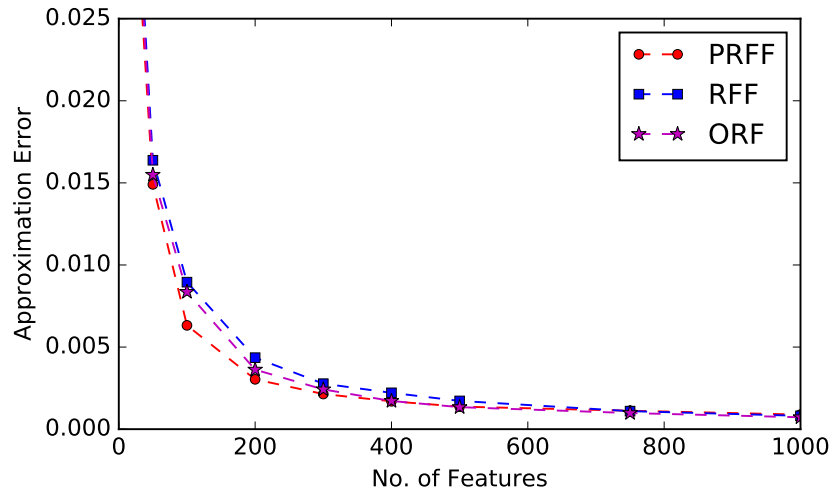

(c) CENSUS

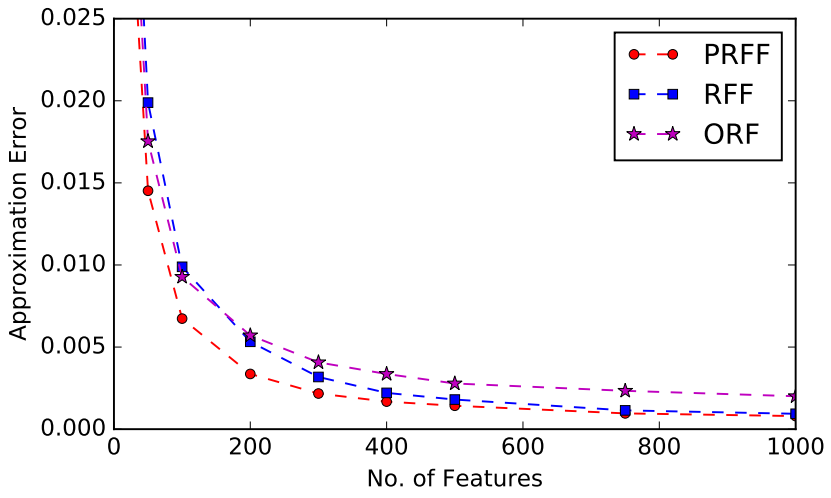

(d) CPU

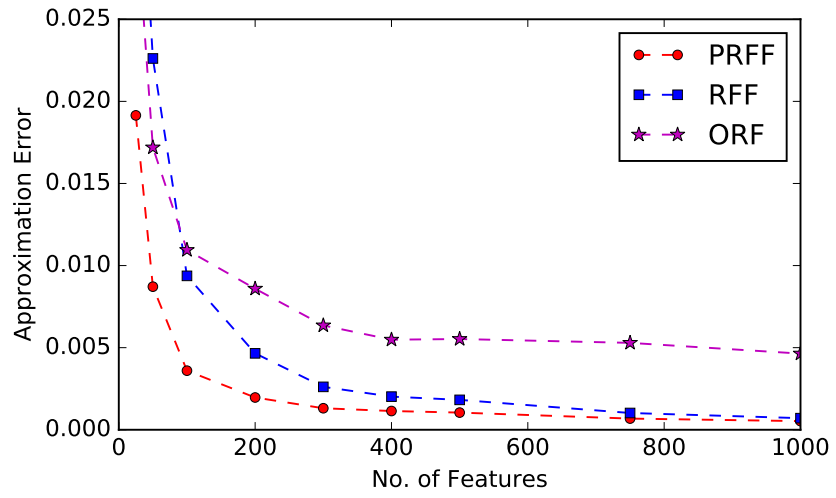

(e) CADATA

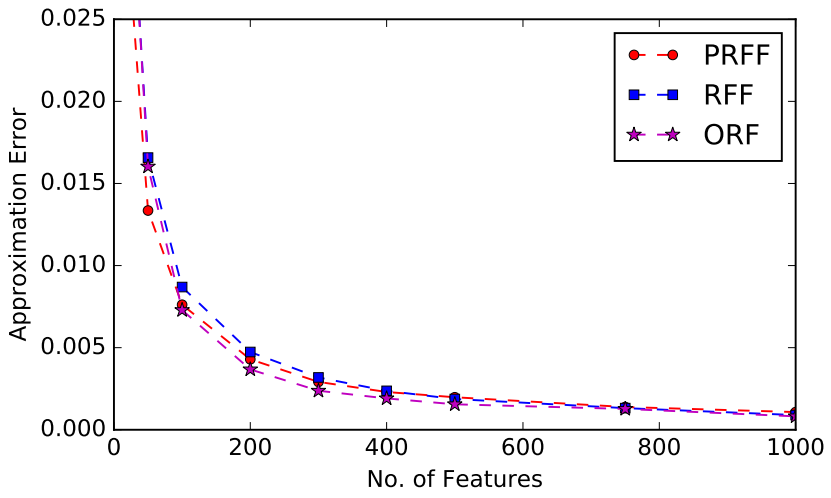

(f) YEARMSD

Fig. 2: Kernel approximation error for the real datasets

data and use it as validation data. We normalized data using standard normal normalization based on the parameters estimated from the training samples before the experiments. The kernel parameter $\sigma$ of the Gaussian kernel was estimated based on $5^{th}$ percentile of pair wise distance between the validation samples. The PRFF coefficients were obtained by minimizing the loss function on the training samples, and the mini-batch size was set to $B = 128$. We experimented with different learning rate $\alpha \in \{50, 25, 10, 5, 1, 0.5, 0.1, 0.01, 0.001\}$, and with different regularization parameter $\lambda \in \{0.1, 0.05, 0.01, 0.001, 0.0001, 0.00001, 0.000001\}$ and only the best results are reported. We set the feature dimension $M = 25, 50, 100, 200, 300, 400, 500$ for all the datasets. The performance of these methods were evaluated by the classification accuracy and mean square error for the classification and regression problems. The classification and regression experiments were performed using ridge regression. We have used the Ridgeclassifier

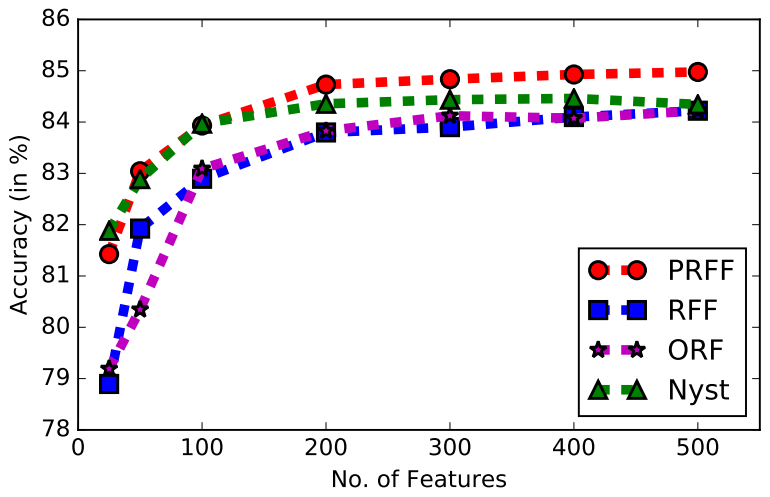

(a) ADULT

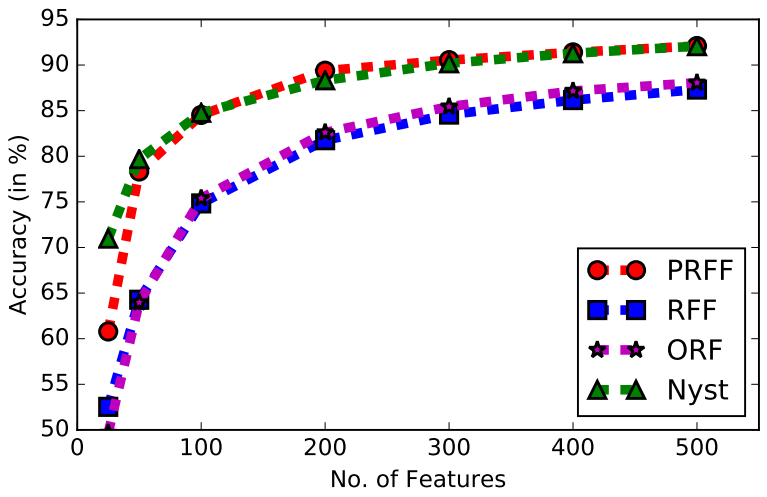

(b) MNIST

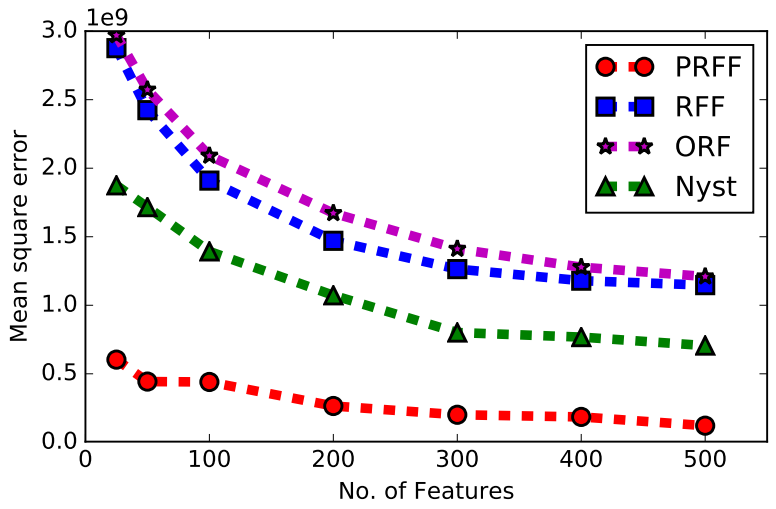

(c) CENSUS

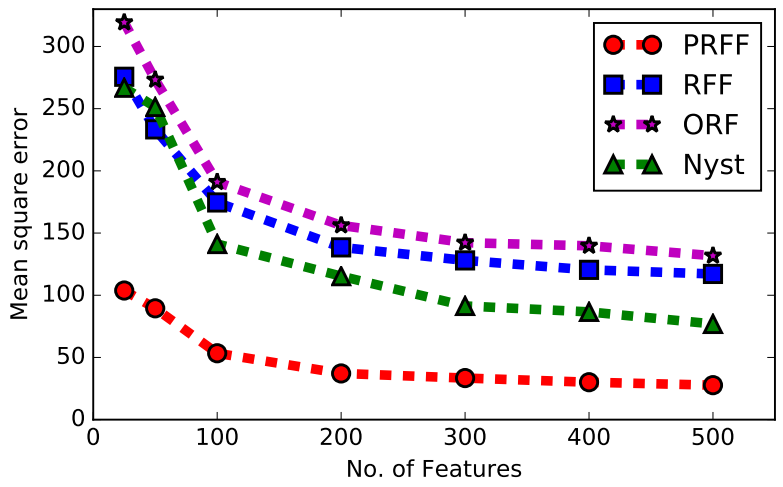

(d) CPU

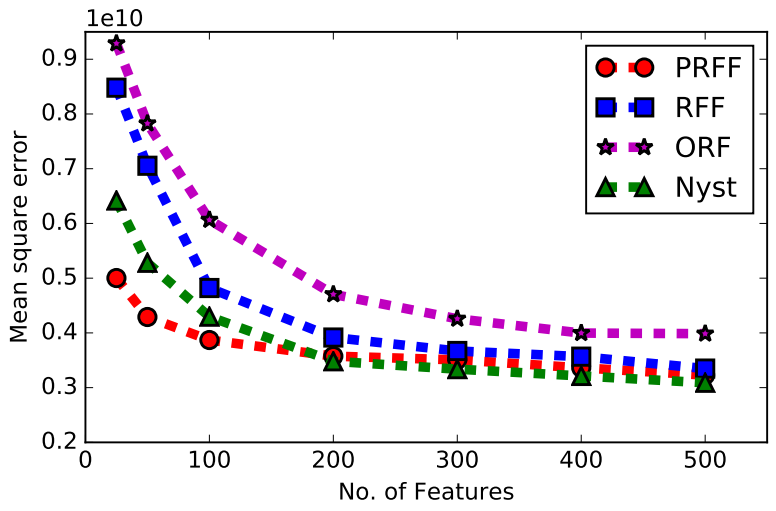

(e) CADATA

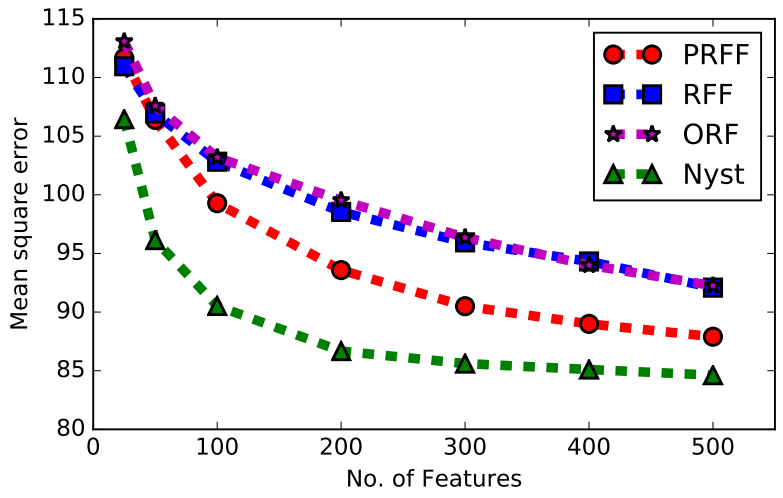

(f) YEARMSD

Fig. 3: Comparison of classification accuracy and mean square error using ridge regression for the classification and regression datasets

module and Ridge module with 'cholesky' solver for the classification and regression problem respectively from the scikit-learn library[7]. The regularization parameter of ridge regression was tuned based on five fold cross validation in the range of $2^C, \quad C = [-15, 1]$ on the validation samples.

Figure 3 shows the results on the real datasets. Our proposed method outperformed the baseline approaches RFF and ORF over all the datasets and it is better or similar to Nyström approach for all but YEARMSD dataset. Interestingly, our method significantly outperformed Nyström approach for the CENSUS and CPU datasets. From the Fig. 3, we can see that the PRFF achieved a performance similar to RFF and ORF with less features. In contrary to the approximation performance, PRFF also obtained better performance than RFF with large number of features. Furthermore, the performance of ORF is slightly better than RFF for the classification problems. This observation is in line with the experiments in [10]. ORF is inferior than RFF for the regression problems. Though ORF has a better kernel approximation error than RFF with few features, this behaviour did not lead to a better prediction performance. This observation also holds for the Nyström approach. In [9], they showed that learning all the coefficients together in the loss function only improve the prediction performance marginally compared to RFF, whereas our proposed strategy clearly showed the improved performance. The superior performance of Nyström approach for the YEARMSD dataset could be because of the large eigen gap, as the Nyström approach has a very low approximation error with less number of features compared to other datasets.

## 5 Conclusion

In this article, we introduced a new method called Pseudo Random Fourier Features (PRRF) to approximate the gram matrix in a data dependent fashion. This approximation has the expressive power of data independent sampling approaches using minimal number of features. The proposed approach incrementally learns each feature dimension. Each new feature dimension is learnt using a batch stochastic gradient descent approach. Experiments using synthetic and real data showed that our proposed method yielded better approximation performance and prediction performance than random Fourier features and orthogonal random features, and comparable or higher performance than Nyström approach.

## Acknowledgments

This work was supported by the French Agence Nationale de la Recherche (ANR) under reference ANR-13-JS02-0005-01 (Asterix project) and by the People Programme (Marie Curie Actions) of the European Unions Seventh Framework

[7] `http://scikit-learn.org/stable/modules/classes.html#module-sklearn.linear_model`

Programme (FP7/2007-2013) under REA Grant PCOFUND-GA-2013-609102, through the PRESTIGE programme coordinated by Campus France.